\documentclass[10pt,twocolumn,letterpaper]{article}

\usepackage[table]{xcolor}
\usepackage{wacv}
\usepackage{tabu}
\usepackage{times}
\usepackage{epsfig}
\usepackage{graphicx}
\usepackage{amsmath}
\usepackage{amssymb}
\usepackage{epsfig}
\usepackage{graphicx}
\usepackage{amsmath}
\usepackage{amssymb}
\usepackage{subcaption}
\usepackage{float}
\usepackage{flushend}

\usepackage{array,setspace, amsfonts,url,bm, cellspace}
\usepackage{epstopdf}
\usepackage{pifont}
\usepackage{multirow, hhline}
\usepackage[flushleft]{threeparttable}

\DeclareMathAlphabet\mathbfcal{OMS}{cmsy}{b}{n}

\newcolumntype{M}[1]{>{\centering\arraybackslash}m{#1}}
\newcommand{\ver}[1]{{\color{red} #1 }}  
\newcommand\notsotiny{\@setfontsize\notsotiny{6}{7}}
\def\@fnsymbol#1{\ensuremath{\ifcase#1\or *\or \dagger\or \ddagger\or
   \mathsection\or \mathparagraph\or \|\or **\or \dagger\dagger
   \or \ddagger\ddagger \else\@ctrerr\fi}}

\def\wacvPaperID{0433} 

\wacvfinalcopy 

\ifwacvfinal
\fi

\ifwacvfinal
\usepackage[breaklinks=true,bookmarks=false,colorlinks]{hyperref}
\else
\usepackage[pagebackref=true,breaklinks=true,colorlinks,bookmarks=false]{hyperref}
\fi

\begin{document}

\title{Generative Adversarial Graph Convolutional Networks for Human Action Synthesis}

\author{Bruno Degardin$^{1,4,5}$, João Neves$^{2,4}$, Vasco Lopes$^{2,4,5}$, João Brito$^5$, Ehsan Yaghoubi$^{3}$, Hugo Proen\c{c}a$^{1,4}$\\
$^1$IT - Instituto de Telecomunica\c{c}\~{o}es, $^2$NOVA LINCS, $^3$C4-Cloud Computing Competence Center\\
 $^4$Universidade da Beira Interior, Portugal \quad\quad $^5$DeepNeuronic \\
{\tt\small bruno.degardin@ubi.pt}
}

\maketitle

\ifwacvfinal
\thispagestyle{empty}
\fi

\begin{abstract}
Synthesising the spatial and temporal dynamics of the human body skeleton remains a challenging task, not only in terms of the quality of the generated shapes, but also of their diversity, particularly to synthesise realistic body movements of a specific action (action conditioning). In this paper, we propose Kinetic-GAN, a novel architecture that leverages the benefits of Generative Adversarial Networks and Graph Convolutional Networks to synthesise the kinetics of the human body. The proposed adversarial architecture can condition up to 120 different actions over local and global body movements while improving sample quality and diversity through latent space disentanglement and stochastic variations. Our experiments were carried out in three well-known datasets, where Kinetic-GAN notably surpasses the state-of-the-art methods in terms of distribution quality metrics while having the ability to synthesise more than one order of magnitude regarding the number of different actions. Our code and models are publicly available at \url{https://github.com/DegardinBruno/Kinetic-GAN}.
\end{abstract}

\vspace{-1em}
\section{Introduction}
\label{sec:intro}
Human behaviour analysis through skeleton-based data has been widely investigated for decades. The advent of deep learning-based architectures increased its popularity even more, mainly due to the robustness of skeleton data in handling dynamic circumstances, appearance variations, and cluttered backgrounds. Over the last decade, the rise of data-driven approaches highly correlates performance with the scale of the learning set. Hence, generating high-quality synthetic human actions can address the problem of limited data. However, existing methods are still severely limited, particularly in conditioning desirable actions and considering the generation at the global movement level. \par

\begin{figure}[t]
  \begin{center}
  \includegraphics[width=8.4cm]{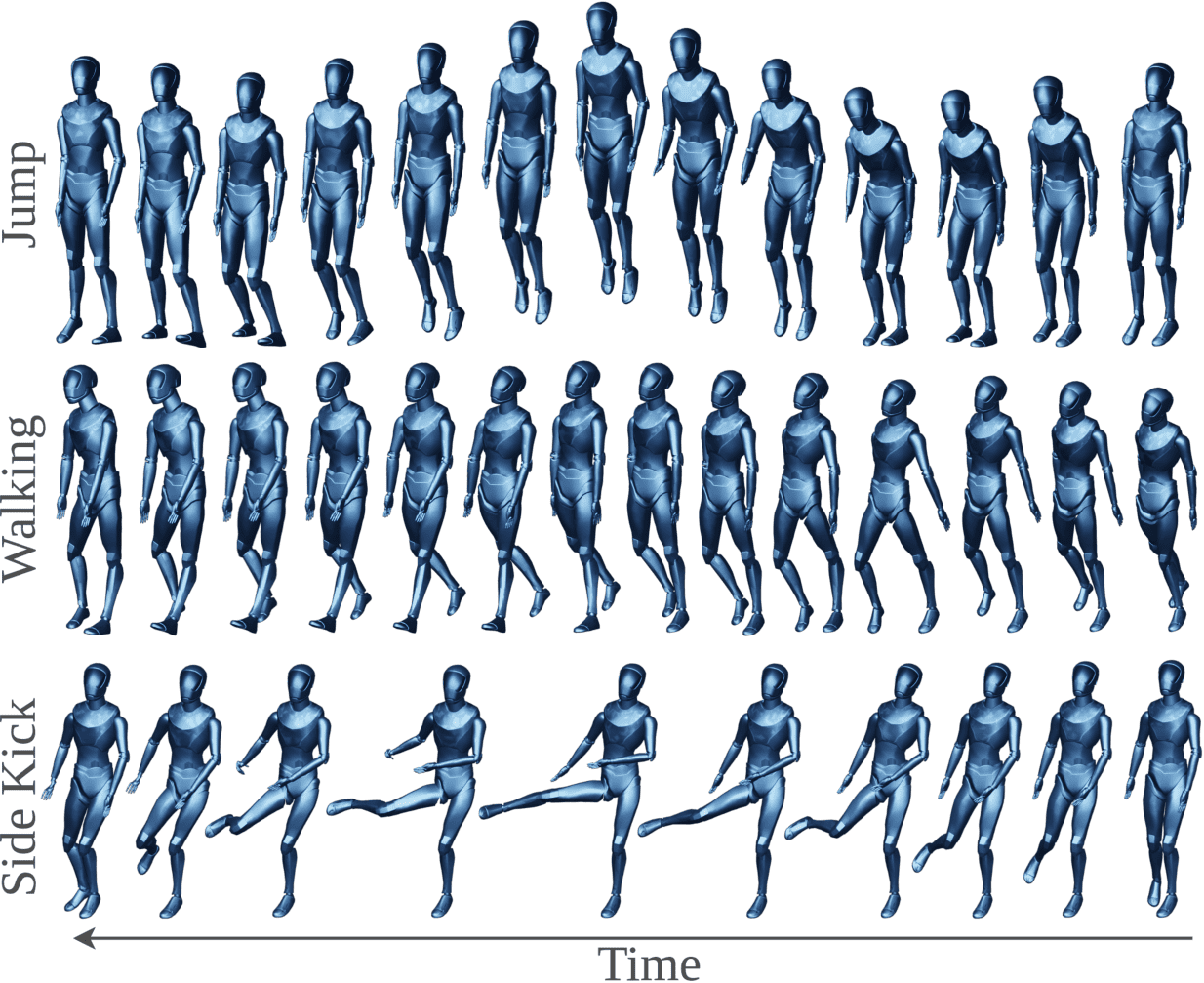}
  \caption{\textbf{Synthetic set of actions} generated by our graph convolutional generator trained on NTU RGB+D \cite{shahroudy2016ntu} (first two rows) and NTU-120 RGB+D \cite{liu2019ntu} (last row). Kinetic-GAN is able to generate up to 120 different actions even under global movement settings. See accompanying video.}
  \label{fig:intro}
  \end{center}
  \vspace{-2em}
\end{figure}

\begin{figure*}[h]
  \begin{center}
  \includegraphics[width=\textwidth]{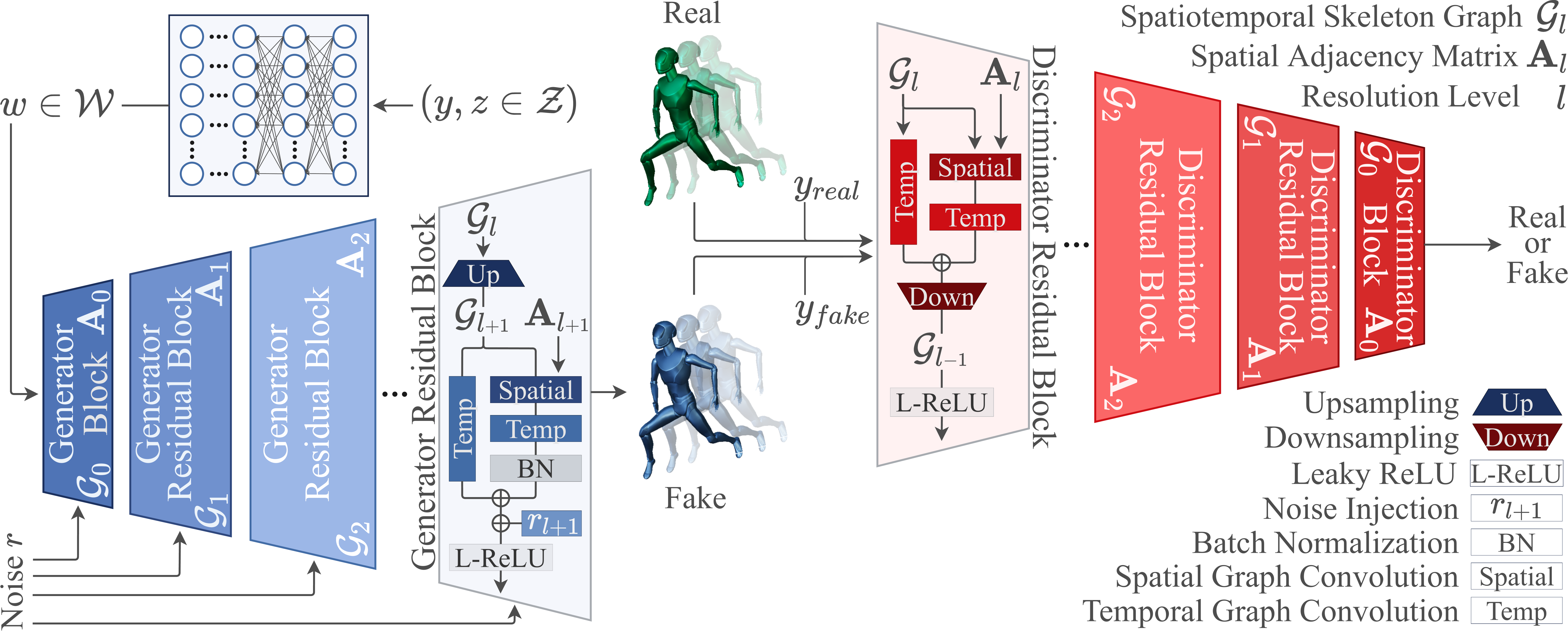}
  \caption{\textbf{Cohesive view of the proposed Kinetic-GAN}. The blue and red components belong to the generator and discriminator, respectively. First, both the Gaussian random noise $\bm{z}$ and the embedded action class representation $\bm{y}$ are concatenated and mapped to an intermediate latent space $\mathbfcal{W}$, which is fed to the generator and upsampled spatially and temporally through each resolution level $l$. The discriminator receives a skeleton graph sequence $\mathbfcal{G}_l$ together with its embedded class (channel-wise concatenation) and learns to discriminate by downsampling spatially (coarsening) and temporally from level $l$ to 0.}
  \label{fig:gan_overview}
  \end{center}
  \vspace{-1.8em}
\end{figure*}

The existing skeleton-based human action synthesis algorithms are classified into two categories: autoregressive and generative approaches. Autoregressive approaches \cite{fragkiadaki2015recurrent, kundu2019unsupervised, zhou2018auto} are usually based on Recurrent Neural Networks (RNNs) and consider the skeleton data as a vector sequence to model an action from several seen frames. Despite the decent quality of the individual samples, autoregressive approaches have two disadvantages: (1) Extracting inherent structural body information using vectorized skeleton sequences is suboptimal. (2) The use of LSTMs \cite{hochreiter1997long}, GRU \cite{bahdanau2014neural}, and Seq2Seq \cite{sutskever2014sequence} potentially limits the scalability regarding the \textit{bidirectional temporal dependency}, i.e., a future frame modifying the past ones, and the increased difficulty of a past frame conditioning a distant future frame. To solve the latter drawback, generative approaches such as \cite{barsoum2018hp, cai2018deep, yan2019convolutional, yu2020structure} use the concept of Generative Adversarial Networks \cite{goodfellow2014generative} (GANs) to produce an entire body skeleton sequence from a latent space. However, this category also has some shortcomings: (1) most approaches still employ manually structured vector sequences to model skeleton data, and (2) they rely on autoregressive techniques and Gaussian processes to solve long-term relationships over the latent space, which greatly limits their scalability in action conditioning.\par

This paper proposes a Generative Adversarial Graph Convolutional Network (Kinetic-GAN) to address the above-mentioned limitations. Our architecture leverages the benefits of GANs and Graph Convolutional Networks (GCNs), such that we generate conditioned human action sequences directly from the latent space while maintaining the long-term relationships between frames. \par
Inspired by the generalization of convolutions from images to graphs, we use spatiotemporal graph convolutions in the generator and discriminator to model skeleton data and exploit the skeleton's inherent graph structure, rather than manually structuring them as coordinate vector sequences. Also, the proposed approach improves the state-of-the-art in action conditioning, controlling up to 120 different actions (examples in Fig. \ref{fig:intro}), whereas previous methods could only control around 10 actions \cite{ habibie2017recurrent, wang2020learning, yu2020structure}. Additionally, the generation of human actions directly from the latent space may inspire some future directions in human behaviour analysis, such as interpretable latent space directions \cite{shen2020interpreting,voynov2020unsupervised} or style transfer between samples \cite{huang2017arbitrary, karras2019style, karras2020analyzing}. Fig. \ref{fig:gan_overview} illustrates the overview of the proposed framework.\par
In summary, our main contributions are three-fold: 1) A new scalable Generative Adversarial Graph Convolutional Network architecture to synthesise human actions. 2) An architecture that can be extended to a conditional model, generating desirable actions up to 120 different classes. 3) We perform extensive experiments on three datasets, NTU RGB+D \cite{shahroudy2016ntu}, NTU-120 RGB+D  \cite{liu2019ntu} and Human$3.6$M \cite{ionescu2013human3}, where Kinetic-GAN exceeds the state-of-the-art performance by a significant margin.

\section{Related Work}
\label{sec:rw}
Human action synthesis regards the generation of understandable spatial and temporal kinematics of the human body skeleton. Current methods extract structural and dynamic patterns from either manually structured sequences or graph-based structures. Those representations are then used to synthesise actions via either autoregressive (to learn temporal dependencies) or generative models (to learn a probability distribution).

\subsection{Skeleton-based Behaviour Analysis}
\label{ssec:rw_skl_intro}
Body pose estimation is one of the auspicious cues in human behaviour analysis. This semantically rich and very descriptive representation of human dynamics attenuates appearance noises that RGB and depth data contain, driving the learning process solely over human behaviour.\par
Over the last decade, skeleton-based behaviour analysis has evolved from pseudo-images with CNNs \cite{ke2017new, li2018co, liu2017enhanced, soo2017interpretable} and sequence coordinate vectors with RNNs \cite{du2015hierarchical, liu2016spatio, liu2017global, song2017end, zhang2017view}, to the solid improvements of GCNs \cite{chen2021multi, cheng2020skeleton, shi2019two, yan2018spatial, zhang2020context, zhang2020semantics} which models skeleton data as a spatiotemporal graph, which better represents the embedded structural information. Still, most current methods \cite{ wang2020learning, yu2020structure} employ manually structured sequence coordinate vectors.

\subsection{Autoregressive Models}
\label{ssec:rw_reg}
Inspired by action prediction models, some works \cite{fragkiadaki2015recurrent, zhou2018auto} employ autoregressive algorithms to generate human actions from several seen frames. Fragkiadaki \etal~\cite{fragkiadaki2015recurrent} proposed to incorporate an encoder-decoder network pre- and post-LSTM-units, capturing the temporal dependencies directly from the low-dimensional representation of the input skeleton. Zhou \etal~\cite{zhou2018auto} presented a conditioned LSTM network, in which the generated data were conditioned at regular intervals of the sequence.

\subsection{Generative Models}
\label{ssec:rw_gen}
Generative adversarial networks \cite{goodfellow2014generative} inspired Generative-based human action synthesis methods \cite{kundu2019unsupervised, wang2020learning, yan2019convolutional, yu2020structure}, whereas some approaches are yet attached to autoregressive techniques \cite{kundu2019unsupervised, wang2020learning, yu2020structure} or Gaussian processes \cite{yan2019convolutional}, which limits their temporal flexibility, stochastic variation, and action conditioning. Kundu \etal~\cite{kundu2019unsupervised} proposed a hierarchical feature fusion based on an RNN auto-encoder architecture with a manually structured tree of limb connections.  Yu \etal~\cite{yu2020structure} suggested using a graph convolutional network on top of an RNN to analyze the latent temporal dependencies. However, this architecture has a limitation regarding temporal length (50 frames) and action conditioning (10 actions). Yan \etal~\cite{yan2019convolutional} proposed a Gaussian process prior to the generator to analyze the dimensions of the latent space one by one (1024 dimensions). Despite reporting temporal long-term latent relations, its heavy computation on latent points correlation limits their diversity and action conditioning. This paper proposes a novel approach that generates long-term human actions without handcrafted procedures (sequence coordinate vectors or Gaussian process priors), enabling us to synthesise far more different actions with significantly more quality even with higher temporal lengths.

\section{Proposed Method}
\label{sec:proposed}
Kinetic-GAN aims to improve the generation quality of controllable action samples while increasing intra-action diversity through stochastic variation and latent space disentanglement. This section presents the proposed approach by initially introducing the GCN adopted, followed by network description, and the discussion of the improvements caried out in the generator, discriminator, and adversarial loss.

\subsection{Graph Convolutional Networks Preliminaries}
Similar to image modelling with GANs, the proposed solution employs a generator and discriminator to perform upsampling and downsampling on samples, respectively. However, since we are modelling skeleton human actions, those operations can not be treated as conventional images with CNNs, which would lead to spatial and temporal distortions. Specifically, the conventional convolutional kernels will lose some structural information embedded in the skeleton data since adjacent joints in the pseudo-image are considered as connected joints. Hence, in each skeleton graph's resolution level $l$, the Kinetic-GAN employs graph convolutions to circumvent this issue due to their ability to exploit the skeleton's inherent graph structure.\par

\begin{figure}[t]
  \begin{center}
  \includegraphics[width=8.4cm]{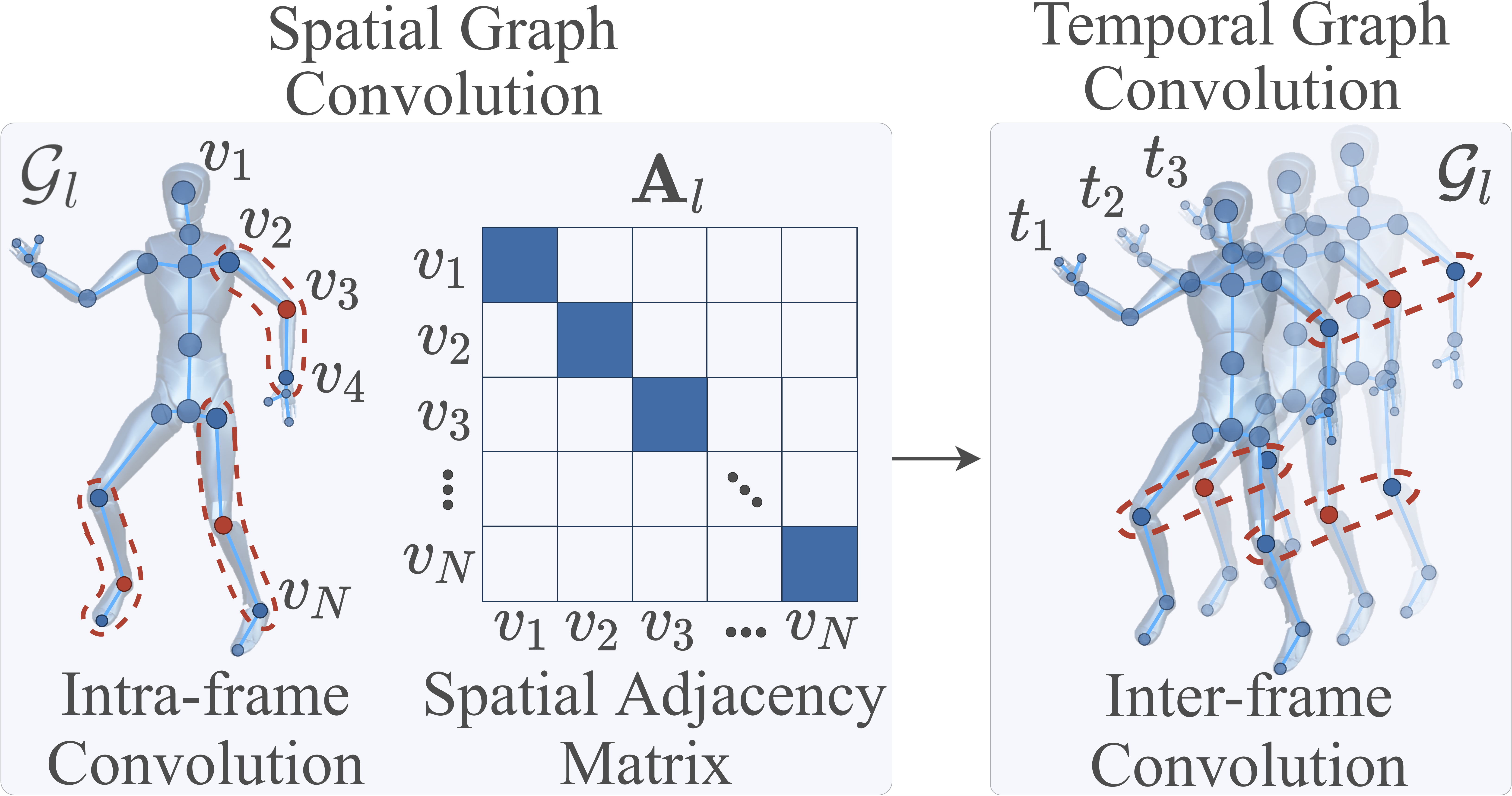}
  \caption{\textbf{Spatiotemporal graph convolution used in Kinetic-GAN}. The spatial graph convolution takes as input a skeleton graph $\mathcal{G}_l$ where its corresponding adjacency matrix $\textbf{A}_l$ handles the intra-frame (spatial) convolution (red dotted line) through the respective root node (red joint) neighbourhood. The temporal (inter-frame) convolution consists of a 1-dimensional convolution performed on the same positional joints across consecutive frames.}
  \label{fig:st-gcn}
  \end{center}
  \vspace{-1.8em}
\end{figure}

In graph convolutional networks (GCNs), a spatiotemporal graph $\mathbfcal{G}_l = (\mathbfcal{V}_l, \mathbfcal{E}_l)$ represents the skeleton data with $N_l$ joints and $T_l$ frames, where $l=\{1, ..., L\}$,  and $L$ is the number of levels of the skeleton graph resolution. Therefore, the feature map of the skeleton sequence is represented as $\textbf{X}_l \in \mathbb{R}^{N_l \times T_l \times C}$, where $C$ is the number of channels, representing the joints coordinates at resolution level $L$. A GCN consists of both spatial and temporal graph convolutions. Typically, in the spatial dimension, an adjacency matrix $\textbf{A}_l \in \{0,1\}^{N_l \times N_l}$ and the corresponding identity matrix $\textbf{I}_l$ define the intra-body joints connections, which are used to regulate the receptive fields of the convolution. Due to its high-level formulation, a partitioning strategy is defined to represent the neighbours set of each joint for constructing convolution operations, whereas $\textbf{A}_l$ and $\textbf{I}_l$ are dismantled into three partitions $p$ (spatial configuration proposed by \cite{yan2018spatial}), so $\textbf{A}_l + \textbf{I}_l = \sum_p$ $\textbf{A}_{l_p}$. For a single frame at resolution level $l$, the graph convolution can be visualized in Fig. \ref{fig:st-gcn} (left), which is computed as:

\begin{equation}\label{eq:1}
   \mathcal{S}(\textbf{X}_l) = \sum ^{p} _{i=1}\mathbf{\Lambda}_{l_i}^{-\frac{1}{2}}\textbf{A}_{l_i}\mathbf{\Lambda}_{l_i}^{-\frac{1}{2}}\textbf{X}_l\textbf{W}_{l_i},
\end{equation}
where the degree matrix $\mathbf{\Lambda}_{l_p}^{ii} = \sum _j (\mathbf{A}_{l_p}^{ij})$ normalizes the adjacency matrix $\mathbf{A}_{l_p}$ through the number of edges attached to each joint node. $\textbf{W}_{l_p}$ denotes the stacked weight vectors for each partition group $p$ from resolution level $l$.\par

Since multiple graph convolutional layers are used, different layers may contain multilevel semantic information \cite{cheng2020skeleton, shi2019two, yan2018spatial}, and simply using $\textbf{A}_l$ in Eq.~\ref{eq:1} results in the same pre-defined spatial weight configuration to every layer. Hence, we also resort to a learnable weight matrix $\textbf{M}_l \in \mathbb{R}^{N_l \times N_l}$ (initialized as an all-one matrix) on each layer of both generator and discriminator. Thus, we can adaptively learn to optimize the spatial weight configuration of $\textbf{A}_l$, and Eq. \ref{eq:1} becomes:

\begin{equation}\label{eq:2}
   \mathcal{S}(\textbf{X}_l) = \sum ^{p} _{i=1}\mathbf{\Lambda}_{l_i}^{-\frac{1}{2}}(\textbf{A}_{l_i} \odot \textbf{M}_l)\mathbf{\Lambda}_{l_i}^{-\frac{1}{2}}\textbf{X}_l\textbf{W}_{l_i},
\end{equation}

Over the temporal axis, considering that consecutive frames define consecutive skeletons, one-dimensional kernels are used as the temporal graph convolution, which is applied after the spatial graph convolution (Fig. \ref{fig:st-gcn}). Finally, our spatiotemporal graph convolution at resolution $l$ is given by convolving the positional features joint-wise as:

\begin{equation}\label{eq:3}
   \mathcal{T}\big(\mathcal{S}(\textbf{X}_l)\big)= \mathcal{S}(\textbf{X}_l) * \textbf{w}_l,
\end{equation}
where $\textbf{w}_l \in \mathbb{R}^{1\times t \times C}$ is the temporal kernel at resolution $l$ with $t$ as the number of frames to be convolved in the kernel.

\subsection{Generative Adversarial Graph Convolutional Network}

\begin{figure}[t]
  \begin{center}
  \includegraphics[width=8.4cm]{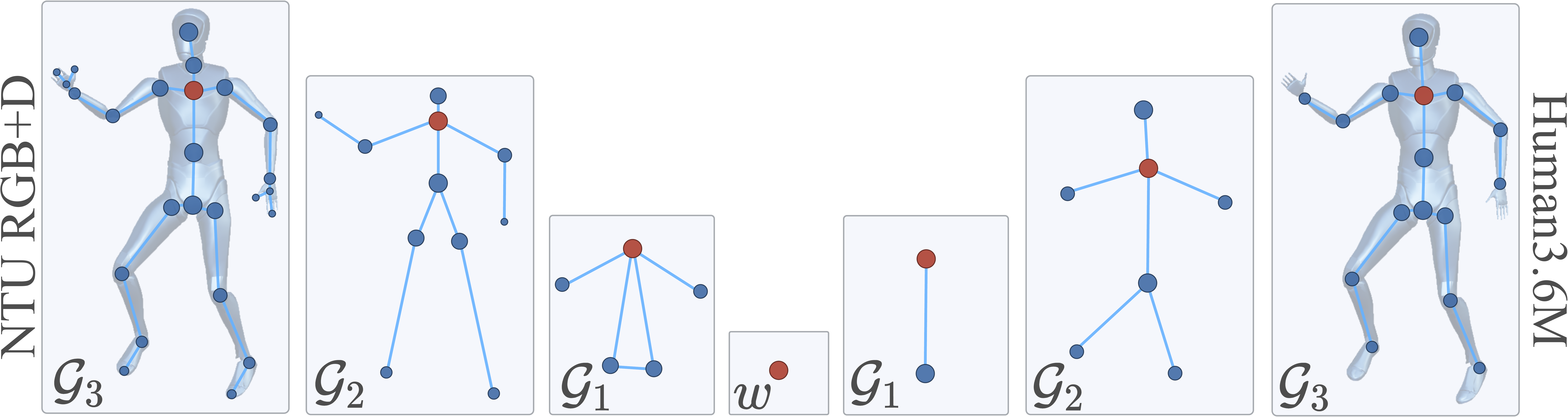}
  \caption{\textbf{Graph upsampling and downsampling paths}. Graph pyramids applied by Kinetic-GAN, where the left side refers to NTU RGB+D \cite{shahroudy2016ntu} and NTU-120 RGB+D \cite{liu2019ntu}, and the right side shows the Human$3.6$M \cite{ionescu2013human3}. The red nodes represents the root node of the respective dataset.}
  \label{fig:updown}
  \end{center}
  \vspace{-1.5em}
\end{figure}

Kinetic-GAN consists of a generative adversarial network composed of a graph convolutional-based generator and discriminator. As previously stated, we perform upsampling and downsampling on samples, respectively, in the generator and discriminator (see Fig. \ref{fig:updown}). Since each spatial graph resolution $\mathcal{G}_l$ has its corresponding adjacency matrix $\textbf{A}_l$, we compute Eq. \ref{eq:3} at any resolution level $l$ of both upsampling and downsampling streams.\par

Typically, generative-based action synthesis methods feed a latent vector to the generator and apply auto-regressive techniques \cite{wang2020learning, yu2020structure}, or Gaussian processes \cite{yan2019convolutional} to capture the temporal relationships in a latent sequence. From a generative perspective, the multiplicative interactions across the latent sequence provoke the entangling of factors of variation \cite{chen2018isolating, desjardins2012disentangling}. We depart from this constraint and propose to synthesise human actions from a single latent point $\bm{z}$ as conventional image generators. However, considering the numerous variation factors in human actions, our generator starts with a non-linear mapping network to transform the latent code $\bm{z} \in \bm{\mathcal{Z}}$ to produce an intermediate latent space $\bm{\mathcal{W}}$. The rationale is to allow less entangled latent factors, as previously confirmed on image modelling \cite{karras2019style, karras2020analyzing, shen2020interpreting}, and consequently increasing the linearity of factors of variation, where the generation of realistic actions becomes easier for the generator. This phenomenon is verified in the ablation study.\par
The proposed generator will gradually increase the resolution of a single intermediate latent point $\bm{w}$ over the spatial and temporal dimensions. First, spatial graph upsampling is computed by introducing new vertices and assigning the corresponding connected joints' average values. One-dimensional interpolation is performed over the temporal axis to increase the number of frames in the skeleton sequence. The discriminator will distinguish between the synthesised actions from the real ones by gradually coarsening the input skeleton's graph, where the vertices are removed, and corresponding neighbours are reconnected. Following the upsampling and downsampling streams from Fig. \ref{fig:updown}, the resolution is increased and decreased as:

\begin{equation}\label{eq:4}
   \textbf{X}_{l+1} = \mathcal{T}\Big( \mathcal{S}\big( Up( \textbf{X}_l ) \big) \Big), \quad
   \textbf{X}_{l-1} = Down\Big( \mathcal{T}\big( \mathcal{S}( \textbf{X}_l ) \big) \Big)
\end{equation}

Additionally, we propose to use residual blocks performing a skip connection with solely a temporal graph convolution to learn the temporal mappings more efficiently, which improves training stability and reduces the appearance of artefacts (we verify this phenomenon in the ablation study). Hence, we define our generator and discriminator residual block as:

\begin{equation}\label{eq:5}
    \begin{split}
       \textbf{X}_{l+1} &= \mathcal{T}\Big( \mathcal{S}\big( Up( \textbf{X}_l ) \big) \Big) + \mathcal{T}\big( Up(\textbf{X}_l) \big)\\
       \textbf{X}_{l-1} &= Down\Big( \mathcal{T}\big( \mathcal{S}( \textbf{X}_l ) \big) + \mathcal{T}(\textbf{X}_l) \Big)
    \end{split}
\end{equation}

\subsection{Conditional Adversarial Training}
The proposed architecture can be extended to a conditional model by feeding additional information about factors that we aim to condition. The generation of desired actions is imperative in human action synthesis; thus, we provide the embedded class information of the action $\bm{y}$ to both generator and discriminator. Specifically, in the generator, the embedded class representation $\bm{y}$ is concatenated to the prior input noise $\bm{z}$ before being mapped to the intermediate latent space $\bm{\mathcal{W}}$. The discriminator is fed with the channel-wise concatenation of the skeleton with the embedded class representation $\bm{y}$. In this paper, we rely on the WGAN-GP objective formulation \cite{gulrajani2017improved}, which is conditioned as:

\begin{equation}
\label{eq:6}
\begin{split}
\min_G \max_D \quad &\overbrace{\mathbb{E}_{\bm{x} \sim \mathbb{P}_r} \left[D(\bm{x} | \bm{y})\right] - \mathbb{E}_{\Tilde{\bm{x}} \sim \mathbb{P}_g} \left[D(\Tilde{\bm{x}} | \bm{y})\right]}^{\text{Discriminator loss}}\\
+\lambda \: &\overbrace{\mathbb{E}_{\hat{\bm{x}} \sim \mathbb{P}_{\hat{\bm{x}}}} [\left(\| \nabla_{\hat{\bm{x}}} D\left(\hat{\bm{x}} | \bm{y}\right)\|_2 - 1 \right)^2]}^{\text{Gradient penalty}},
\end{split}
\end{equation}
where $\mathbb{P}_r$ is the data distribution and $\mathbb{P}_g$ is the model distribution implicitly defined by $\Tilde{\bm{x}} = G(\bm{z},\bm{y}), \bm{z} \sim p(\bm{z})$ (the input $\bm{z}$ is sampled from a noise distribution $p$, which is then concatenated with the embedded action class representation $\bm{y}$). $\mathbb{P}_{\hat{\bm{x}}}$ is sampled uniformly along straight lines between pairs of points sampled from the data distribution $\mathbb{P}_r$ and generator distribution $\mathbb{P}_g$. The loss weight $\lambda$ for gradient penalty is set to 10 in all experiments.

\subsection{Improving Quality and Diversity}
\subsubsection{Stochastic Variation}
As previously stated, the current state-of-the-art methods are still attached to autoregressive techniques and Gaussian processes. Such procedures performed over the latent space reduce the ability of variation between samples.\par
Aside from the non-linear mapping network to attain less entangled latent factors, we propose an individual stochastic variation to circumvent this issue without affecting the skeleton structure in the action sequence itself. Specifically, we add random noise to each joint independently in the generator after each spatiotemporal graph convolution, which can be learned by assigning weights to every channel. Our noise injection operation is introduced in the generator's residual block as:

\begin{equation}\label{eq:7}
   \textbf{X}_{l+1} = \mathcal{T}\Big( \mathcal{S}\big( Up( \textbf{X}_l ) \big) \Big) + \mathcal{T}\big( Up(\textbf{X}_l) \big) + \bm{r}_{l+1}\textbf{w}_{l+1},
\end{equation}
where $\bm{r}_{l+1}$ denotes the Gaussian random noise for each joint to be added at resolution level $l+1$, and $\textbf{w}_{l+1}$ is the respective weight vector for each channel. The rationale is to provide a second input to the generator, which is handled to produce variation between samples without forcing the generator to use earlier activations from the latent space to generate random noise. Moreover, each layer has a corresponding per-channel weight $\textbf{w}_{l}$ and receives a new random noise $\bm{r}_{l}$, giving the flexibility needed to adaptively learn to produce stochastic variation without compromising the skeleton structure of the action (this phenomenon is verified in the ablation study).\par

\begin{figure}[t]
  \begin{center}
  \includegraphics[width=8.4cm]{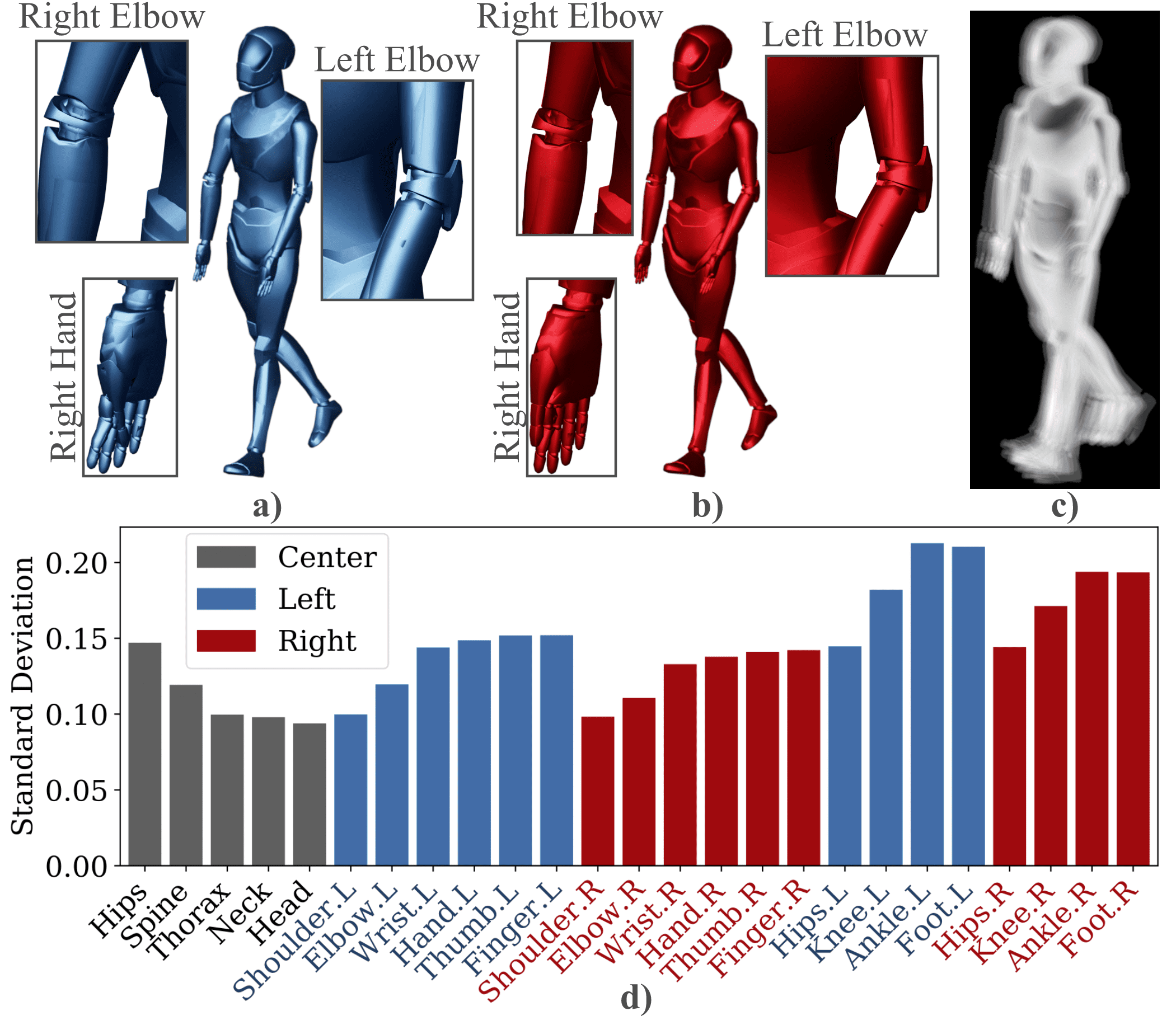}
  \caption{\textbf{Examples of stochastic variation}. a) and b) Two human skeletons at the same frame from two action sequences (walking) generated by the same latent point. Zoom-in areas correspond to the respective coloured skeleton. c) Standard deviation over 100 different realizations from the respective frame, highlighting the skeleton parts affected by the noise. d) Cumulative standard deviation of each joint w.r.t. c). A cohesive behaviour is reproduced, where naturally the edges of limbs (hands, fingers, heels, feet) have a higher deviation than their parent joints.}
  \label{fig:stochastic}
  \end{center}
  \vspace{-1.7em}
\end{figure}

Fig.~\ref{fig:stochastic} illustrates the effect of stochastic variation on a walking sequence generated from the same latent point with different noise realizations. \ref{fig:stochastic} a) and b) show how noise affects different parts of the skeletons at the same frame, and c) highlights the standard deviation from 100 different realizations at the same frame, where most affected areas are the legs and arms (w.r.t. the walking action). The lower plot d) comprises a quantitative view of the standard deviation of each joint in c). It can be seen that the noise injector correctly learned the intrinsic natural variation of limbs since it affects more the edges, such as hands and feet than their corresponding parent joints (knees and elbows). Moreover, left limbs (blue) are slightly higher than their corresponding right ones (red) since the respective frame of the skeleton is moving the left leg forward, so naturally, its deviation will be higher than the right leg. For instance, in a throwing action, the arm that throws the object will always have more variation than the other arm. See accompanying video.

\subsubsection{Reducing Spatial Artefacts}
Typically, the increased quality of generated samples leads to unpleasant artefacts, which is a phenomenon already well-known in image modelling \cite{brock2018large,galteri2017deep,karras2020analyzing,wang2018esrgan}. Those are often the result of normalization methods over the generator, which improves training stability by eliminating covariate shift. Even so, the feature map's normalization omits any information concerning the individual feature's magnitude. It is assumed that the generator magnifies those magnitudes, and through the normalization process, it becomes unnoticed by the discriminator \cite{ karras2020analyzing, wang2018esrgan}.\par
\begin{figure}[h]
  \begin{center}
  \includegraphics[width=8.4cm]{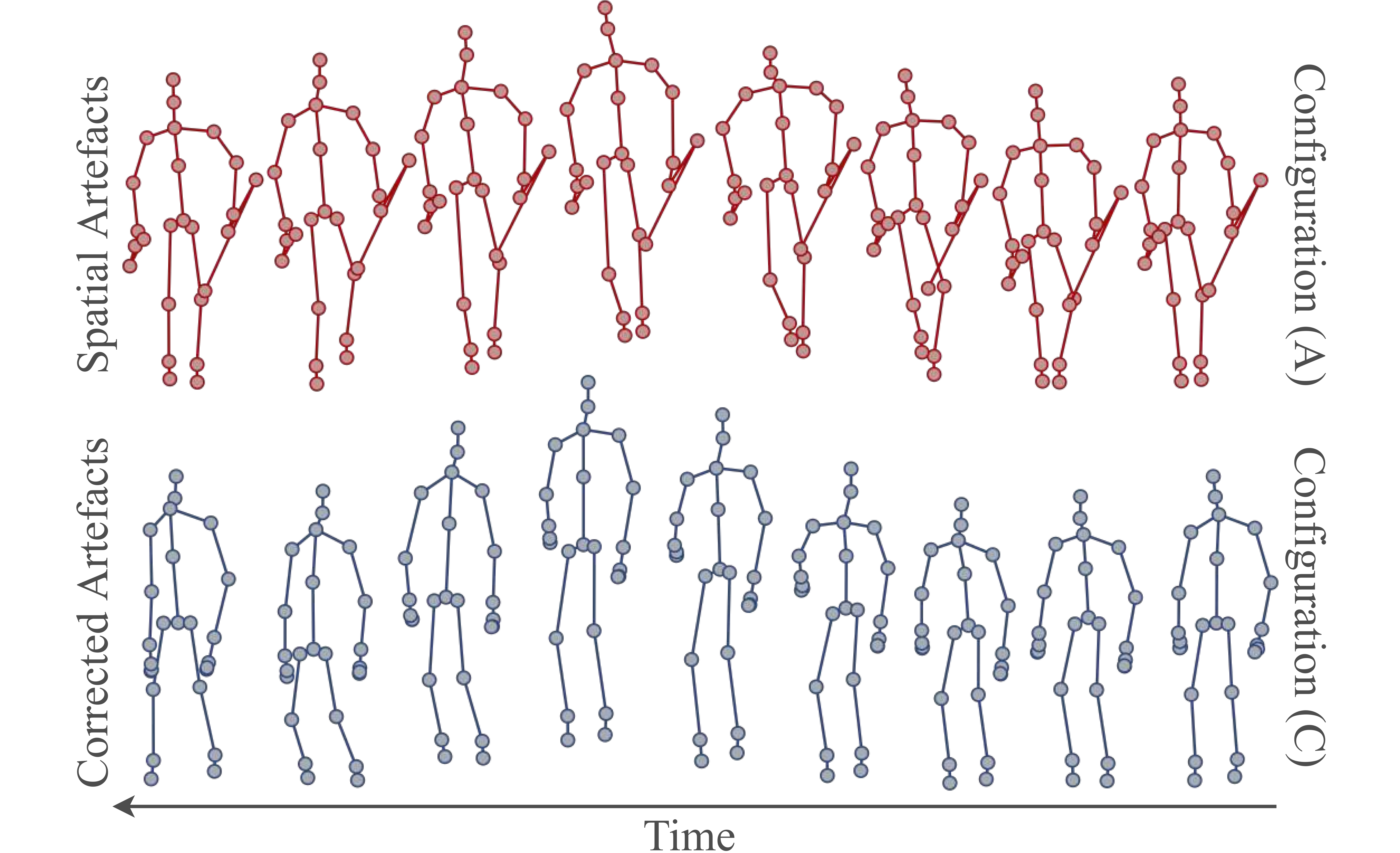}
  \caption{\textbf{Spatial artefacts in action synthesis}. First row shows an occurrence of spatial artefacts on a jumping action sequence generated by configuration {\small{(A)}} in Table \ref{tab:ablation_full}. Similar to image modelling, the remaining information is unaffected as we can identify a human skeleton jumping. Second row shows a jumping action sequence with corrected artefacts by configuration {\small{(C)}} in Table \ref{tab:ablation_full}.}
  \label{fig:artefact}
  \end{center}
  \vspace{-2.5em}
\end{figure}

This hypothesis is supported by the fact that removing these normalizations from our generator eradicated the appearance of artefacts. Still, it disabled our ability to manipulate human actions. Therefore, we found that reducing the number of layers that employ batch normalization led to better stability in action conditioning while completely removing artefacts. Fig. \ref{fig:artefact} illustrates an artefact example in action synthesis.

\subsubsection{Truncation Trick on $\mathbfcal{W}$}
As described, the Kinetic-GAN samples $\bm{z}$ from $\mathcal{N}(0,1)$ and maps it an intermediate representation $\mathbfcal{W}$ of that distribution, which is then fed to the generator. So naturally, ranges of low density (in the training data) are not well represented, becoming difficult to learn for the generator, resulting in an important open problem in generative algorithms.\par
\begin{figure}[t]
  \begin{center}
  \includegraphics[width=8.4cm]{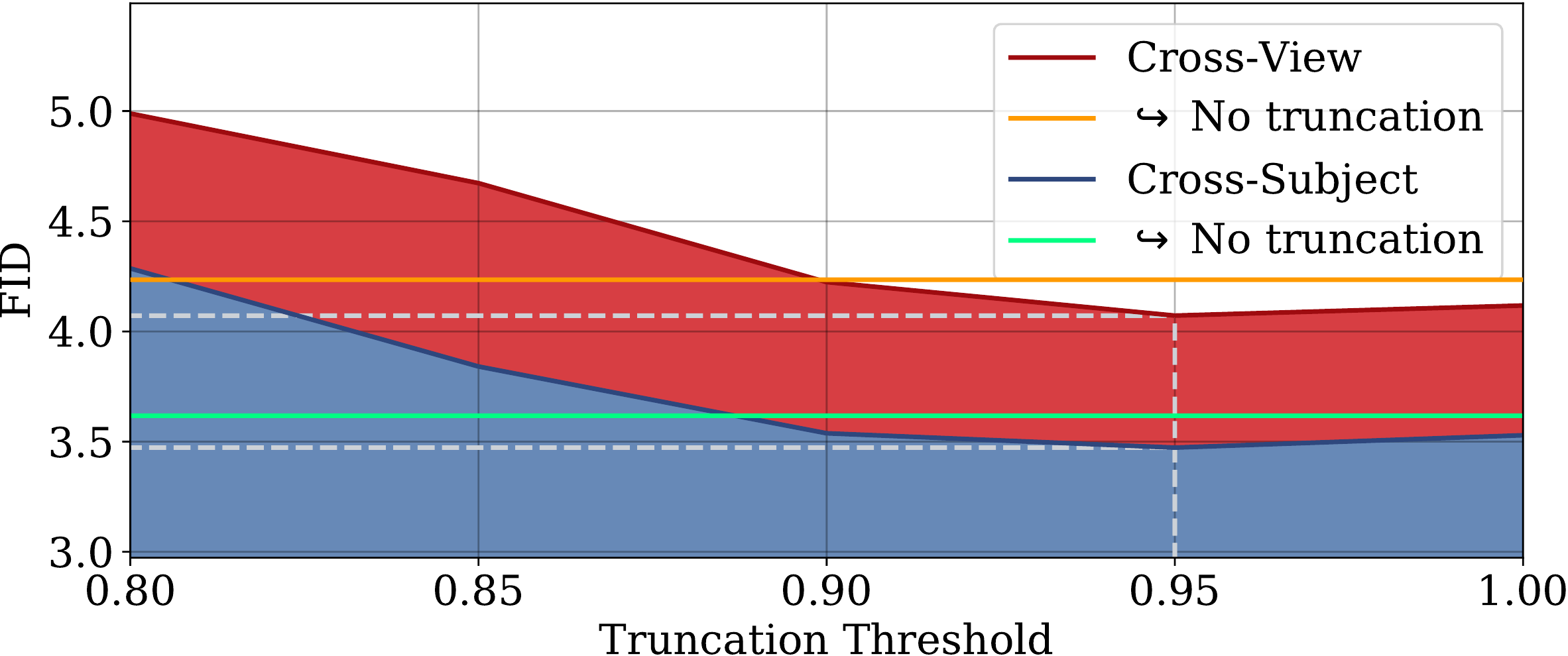}
  \caption{\textbf{Improvements through the ''truncation trick''}. The horizontal lines correspond to the respective FID obtained in Table \ref{tab:ablation_full} without truncations. All action sequences illustrated in this paper uses $\psi = 0.95$.}
  \label{fig:trunc}
  \end{center}
  \vspace{-1.8em}
\end{figure}

As previously confirmed, sample quality can be improved from truncated \cite{brock2018large,karras2019style,marchesi2017megapixel}, or shrunken \cite{kingma2018glow} sampling spaces. Thus, despite some variation losses, we follow the same approach to balance sample quality and diversity. During inference time, we scale the deviation of a given intermediate latent point $\bm{w}$ from the centre mass of $\mathbfcal{W}$ as:

\begin{equation}\label{eq:8}
   \bm{w}' = \mathbb{E}_{\bm{z} \sim \mathbb{P}_{\bm{z}}} \left[f(\bm{z})\right] + \psi(\bm{w} - \mathbb{E}_{\bm{z} \sim \mathbb{P}_{\bm{z}}} \left[f(\bm{z})\right]),
\end{equation}
where $\psi \leq 1$, $f(\cdot)$ denote our mapping network, and $\mathbb{P}_{\bm{z}}$ is the latent space distribution from 1000 points. Despite Brock \etal~\cite{brock2018large} reporting that only a subset of networks is manageable to such truncations even when orthogonal regularization is used, truncation in our intermediate space $\mathbfcal{W}$ successfully works even without adjustments to the loss function. As shown in Fig.~\ref{fig:trunc}, we can increase the generation quality in both benchmarks of NTU RGB+D \cite{shahroudy2016ntu}. However, for  $\psi \leq 0.9$, the variation starts to decrease and, consequently, the FID starts to increase.

\section{Experiments and Discussion}
\label{sec:Results}
In this section, the experiments are split into local and global movement settings concerning if the 3D locations were positionally normalized into 2D space or not. The ablation studies verify the efficacy of the proposed model's properties over global movement. Then, our best performing model is compared to the state-of-the-art approaches over four datasets: NTU RGB+D \cite{shahroudy2016ntu} and NTU-120 RGB+D  \cite{liu2019ntu} (global movement) and Human$3.6$M \cite{ionescu2013human3} and NTU-2D RGB+D \cite{shahroudy2016ntu} (local movement). 

\subsection{Datasets and evaluation metrics}
\textbf{NTU RGB+D} \cite{shahroudy2016ntu}. This dataset is composed of 56,880 video samples with 60 action classes. 3D skeleton data from 40 volunteers are provided for each action sample, with 25 joints for each skeleton. Authors recommend two benchmarks: 1) cross-subject, where models are trained with 20 subjects and tested with the remaining ones; and 2) cross-view, where models are trained with camera views 2 and 3 and tested on camera view 1. A curated version of this dataset is also used (2D joints locations, global to local movement normalization and selected samples) for a fair comparison in Table \ref{tab:local}, denoted by NTU-2D RGB+D. \par

\textbf{NTU-120 RGB+D} \cite{liu2019ntu}. This dataset is an extended version of its predecessor, comprising 114,480 video samples with 120 action classes. Samples were captured with three camera views in 32 different setups and 106 volunteers with 25 body joints for each skeleton. Authors recommend two benchmarks: 1) cross-subject, where models are trained with 53 subjects and tested with the remaining ones, and 2) cross-setup, where models are trained from samples with even setup IDs and tested on odd setup IDs.\par

\textbf{Human$\mathbf{3.6}$M} \cite{ionescu2013human3}. This dataset is a more simplistic set with 2D human motions and 15 body joints for each skeleton. For a fair comparison between the state-of-the-art, the same pre-processing and settings as \cite{wang2020learning, yu2020structure} were followed with corresponding 10 action classes.\par

\textbf{Evaluation metrics}. Two evaluation metrics are used for estimating the quality of synthetic samples. Similar to image modelling, we use the Fréchet Inception Distance (FID) \cite{heusel2017gans}, measuring the distance between the real data distribution and the synthesised one, considering the output activations of a specific layer from an InceptionV3 network \cite{szegedy2016rethinking}. Additionally, we measure the Maximum Mean Discrepancy ($\text{MMD}$) between real and synthetic samples based on a two-sample test to measure the discrepancy of both distributions. The $\text{MMD}$ over motion dynamics corresponds to the average $\text{MMD}$ across each frame, denoted by $\text{MMD}_{a}$, and the MMD over whole sequences are indicated as $\text{MMD}_{s}$. For joints over 2D space, only the $\text{MMD}$ is used.\par

\subsection{Global Movement Settings}
\label{sec:global}

\subsubsection{Ablation study}
\label{ssec:ablation}
Before diving into state-of-the-art comparisons, we first demonstrate experimentally that Kinetic-GAN properties improve sample quality considerably.\par

\newcolumntype{C}[1]{>{\centering\arraybackslash}p{#1}}
\newcolumntype{l}[1]{>{\arraybackslash}p{#1}}
\renewcommand{\arraystretch}{0.9}
\def\textBF#1{\sbox\CBox{#1}\resizebox{\wd\CBox}{\ht\CBox}{\textbf{#1}}}

 
\begin{table}[h]
\begin{center}
\begin{tabular}{|C{0.001\textwidth}|l{0.097\textwidth}|C{0.03\textwidth}|C{0.03\textwidth}|C{0.03\textwidth}|C{0.03\textwidth}|C{0.03\textwidth}|C{0.03\textwidth}|}
\cline{2-8}
\multicolumn{1}{c|}{} & 
\textbf{\footnotesize{Method}} & 
\textbf{\tiny{$\mathbf{FID}$}} & 
\negthinspace\negthinspace\textbf{\tiny{$\mathbf{MMD_{a}}$}} & 
\negthinspace\negthinspace\textbf{\tiny{$\mathbf{MMD_{s}}$}} &
\textbf{\tiny{$\mathbf{FID}$}} &
\negthinspace\negthinspace\textbf{\tiny{$\mathbf{MMD_{a}}$}} & 
\negthinspace\negthinspace\textbf{\tiny{$\mathbf{MMD_{s}}$}} \\
\cline{2-8}
\noalign{\smallskip}
\hline
\parbox[t]{1mm}{\multirow{8}{*}{\hspace{-0.21em}\rotatebox[origin=c]{90}{\footnotesize{\textbf{NTU RGB+D}}}}}
& \multicolumn{1}{c|}{\cellcolor{gray!25}} & 
\multicolumn{3}{c|}{\cellcolor{gray!25}  \footnotesize{\emph{Cross-Subject} }}& 
\multicolumn{3}{c|}{\cellcolor{gray!25}  \footnotesize{\emph{Cross-View} }}\\
&\scriptsize{c-GAN \cite{mirza2014conditional} } & 
\negthinspace\negthinspace\scriptsize{$27.480$}  &
\negthinspace\scriptsize{$0.919$}   &  
\negthinspace\scriptsize{$0.975$} &
\negthinspace\negthinspace\scriptsize{$31.875$}   &
\negthinspace\scriptsize{$0.993$}    &  
\negthinspace\scriptsize{$1.088$} \\
&\scriptsize{Baseline \cite{yan2019convolutional} } & 
\negthinspace\scriptsize{$6.030$}   &  
\negthinspace\scriptsize{$0.873$}    &  
\negthinspace\scriptsize{$0.954$} &
\negthinspace\scriptsize{$7.114$}   &  
\negthinspace\scriptsize{$0.910$}    &  
\negthinspace\scriptsize{$0.991$} \\
\cline{2-8}
&\scriptsize{A \hspace{0.6em} Proposed} & 
\negthinspace\scriptsize{$5.621$}  & 
\negthinspace\scriptsize{$0.836$}&
\negthinspace\scriptsize{$0.927$}   &  
\negthinspace\scriptsize{$6.528$}      &  
\negthinspace\scriptsize{$0.883$}    &  
\negthinspace\scriptsize{$0.953$} \\
&\scriptsize{B + No Residual} & 
\negthinspace\negthinspace\scriptsize{$19.723$}   & 
\negthinspace\scriptsize{$0.892$}     &  
\negthinspace\scriptsize{$0.961$}&
\negthinspace\negthinspace\scriptsize{$21.331$}   &  
\negthinspace\scriptsize{$0.971$}    &  
\negthinspace\scriptsize{$1.030$} \\
&\scriptsize{C + Regular BN} & 
\negthinspace\scriptsize{$4.751$}   &  %
\negthinspace\scriptsize{$0.815$}    &  
\negthinspace\scriptsize{$0.917$}    &  
\negthinspace\scriptsize{$5.328$}    &  
\negthinspace\scriptsize{$0.867$}    &  
\negthinspace\scriptsize{$0.940$} \\
&\scriptsize{D + Noise Inject} & 
\negthinspace\scriptsize{$4.698$}   &  
\negthinspace\scriptsize{$0.811$}&
\negthinspace\scriptsize{$0.895$}   &  
\negthinspace\scriptsize{$5.102$}     &  
\negthinspace\scriptsize{$0.851$}    &  
\negthinspace\scriptsize{$0.933$} \\
&\scriptsize{E + Mapping Net} & 
\negthickspace\scriptsize{$\mathbf{3.618}$} & 
\negthickspace\scriptsize{$\mathbf{0.772}$}  &  
\negthickspace\scriptsize{$\mathbf{0.871}$}    &  
\negthickspace\scriptsize{$\mathbf{4.235}$}     &    
\negthickspace\scriptsize{$\mathbf{0.824}$}    &  
\negthickspace\scriptsize{$\mathbf{0.913}$} \\\hline   

\end{tabular}
\vspace{-1em}
\end{center}
\caption{\textbf{Evaluating different generator designs}. The FID and MMD scores (lower is better) between real and synthetic samples generated under global body movement.}
\label{tab:ablation_full}
\end{table}

\textbf{Different generator designs}. In Table \ref{tab:ablation_full}, we compare the FID and MMD for various generator architectures in both benchmarks of NTU RGB+D, evaluating each distribution under the respective settings of each method. While the baseline (CSGN \cite{yan2019convolutional}) generates local movement in 3D space, we generate global movement and still exceed their performance by a significant margin. In CSGN \cite{yan2019convolutional}, the application of Gaussian processes limits their generation diversity due to the multiplicative interactions across each dimension of the latent space, which emulate the entangling of factors of variation. Additionally, the application of such processes has a high computational cost.\par

We start with our proposed configuration {\small{(A)}} with conditional sampling applying temporal skip connections and batch normalization over each generator layer. We then confirm the importance of residual blocks in our method by removing them from the generator  {\small{(B)}}, which clearly reduces sample quality, mainly due to training instability and, consequently, the appearance of artefacts. Since configuration {\small{(A)}} also generates occasional artefacts and eradicating batch normalization sacrificed our ability to generate desirable actions, we propose to regularize the use of batch normalization {\small{(C)}} by removing it when spatial upsampling is performed. This allows us to eliminate artefacts and have complete control to produce desirable actions while still improving sample quality compared to {\small{(A)}}. We also introduce noise injection {\small{(D)}} that further improves the results by adding even more diversity. Finally, we also include our mapping network {\small{(E)}}, where we distinctly overcome previous baselines and configurations due to the better disentangled latent space. \par


\begin{table}[h]
\begin{center}
\begin{tabular}{|C{0.001\textwidth}|l{0.098\textwidth}|C{0.03\textwidth}|C{0.03\textwidth}|C{0.03\textwidth}|C{0.03\textwidth}|C{0.03\textwidth}|C{0.03\textwidth}|}
\cline{2-8}
\multicolumn{1}{c|}{} & 
\textbf{\footnotesize{Method}} & 
\textbf{\tiny{$\mathbf{FID}$}} & 
\negthinspace\negthinspace\textbf{\tiny{$\mathbf{MMD_{a}}$}} & 
\negthinspace\negthinspace\textbf{\tiny{$\mathbf{MMD_{s}}$}} &
\textbf{\tiny{$\mathbf{FID}$}} &
\negthinspace\negthinspace\textbf{\tiny{$\mathbf{MMD_{a}}$}} & 
\negthinspace\negthinspace\textbf{\tiny{$\mathbf{MMD_{s}}$}} \\
\cline{2-8}
\noalign{\smallskip}
\hline
\parbox[t]{1mm}{\multirow{7}{*}{\hspace{-0.21em}\rotatebox[origin=c]{90}{\footnotesize{\textbf{NTU RGB+D}}}}}
& \multicolumn{1}{c|}{\cellcolor{gray!25}} & 
\multicolumn{3}{c|}{\cellcolor{gray!25}  \footnotesize{\emph{Cross-Subject} }}& 
\multicolumn{3}{c|}{\cellcolor{gray!25}  \footnotesize{\emph{Cross-View} }}\\
&\scriptsize{ D Mapping Net $\emptyset$} & 
\negthinspace\scriptsize{$4.698$}  & 
\negthinspace\scriptsize{$0.811$}&
\negthinspace\scriptsize{$0.895$}   &  
\negthinspace\scriptsize{$5.102$}&    
\negthinspace\scriptsize{$0.851$}    &  
\negthinspace\scriptsize{$0.933$}  \\
\cline{2-8}
&\scriptsize{\hspace{0.67em}  Mapping Net 1} & 
\negthinspace\scriptsize{$4.049$}  & 
\negthinspace\scriptsize{$0.792$}     &  
\negthinspace\scriptsize{$0.892$}&
\negthinspace\scriptsize{$4.792$} &  
\negthinspace\scriptsize{$0.847$}    &  
\negthinspace\scriptsize{$0.931$} \\
&\scriptsize{\hspace{0.67em} Mapping Net 2} & 
\negthinspace\scriptsize{$3.987$}  & 
\negthinspace\scriptsize{$0.780$}     &  
\negthinspace\scriptsize{$0.885$}&
\negthinspace\scriptsize{$4.698$} &  
\negthinspace\scriptsize{$0.839$}    &  
\negthinspace\scriptsize{$0.929$} \\
&\scriptsize{E Mapping Net 4} & 
\negthickspace\scriptsize{$\mathbf{3.618}$}  & 
\negthickspace\scriptsize{$\mathbf{0.772}$}  &  
\negthickspace\scriptsize{$\mathbf{0.871}$}    &  
\negthinspace\scriptsize{$4.473$}  &  
\negthinspace\scriptsize{$0.831$}    &  
\negthinspace\scriptsize{$0.926$} \\
&\scriptsize{E Mapping Net 6} & 
\negthinspace\scriptsize{$3.849$}  & 
\negthinspace\scriptsize{$0.801$}  &  
\negthinspace\scriptsize{$0.889$}    &  
\negthickspace\scriptsize{$\mathbf{4.235}$}  &  
\negthickspace\scriptsize{$\mathbf{0.824}$}  &  
\negthickspace\scriptsize{$\mathbf{0.913}$} \\
&\scriptsize{\hspace{0.67em} Mapping Net 8} & 
\negthinspace\scriptsize{$4.396$}  & 
\negthinspace\scriptsize{$0.805$}     &  
\negthinspace\scriptsize{$0.891$}&
\negthinspace\scriptsize{$4.610$}&  
\negthinspace\scriptsize{$0.837$}    &  
\negthinspace\scriptsize{$0.919$} \\\hline

\end{tabular}
\vspace{-1em}
\end{center}
\caption{\textbf{Importance of the mapping network}. The number in method indicates the depth of the mapping network.}
\label{tab:mapping}
\end{table}

\textbf{Mapping network effectiveness}. The importance of the mapping network is presented in Table \ref{tab:mapping}, where we compare configuration {\small{(D)}} (not using a mapping network) with the increased number of layers in the mapping network. The cross-view benchmark required a deeper network than the cross-subject to attain the optimal performance. We justify this phenomenon due to the increasing number of different subjects in the data, which results in a more complex latent space representation. This indeed confirms the importance of the mapping network since naturally, if there is a greater number of different people, the variations will be higher.

\subsubsection{Synthesising 120 different actions}
\label{ssec:120actions}
Apart from our ablation studies in Section \ref{ssec:ablation} which were produced considering global movement settings, Table \ref{tab:ntu120} also presents the performance obtained by Kinetic-GAN over the NTU-120 RGB+D with 120 different actions, being currently the most extensive and challenging set with 3D joints annotations. Despite achieving a significantly better quality than previous methods, the ability to generate 120 different actions under global movement settings is a substantial improvement in comparison with previous state-of-the-art where they could only generate under local movement settings, and just 10 different actions \cite{habibie2017recurrent, wang2020learning,yu2020structure}.

\begin{table}[h]
\begin{center}
\begin{tabular}{|C{0.001\textwidth}|l{0.097\textwidth}|C{0.03\textwidth}|C{0.03\textwidth}|C{0.03\textwidth}|C{0.03\textwidth}|C{0.03\textwidth}|C{0.03\textwidth}|}
\cline{2-8}
\multicolumn{1}{c|}{} & 
\textbf{\footnotesize{Method}} & 
\textbf{\tiny{$\mathbf{FID}$}} & 
\negthinspace\negthinspace\textbf{\tiny{$\mathbf{MMD_{a}}$}} & 
\negthinspace\negthinspace\textbf{\tiny{$\mathbf{MMD_{s}}$}} &
\textbf{\tiny{$\mathbf{FID}$}} &
\negthinspace\negthinspace\textbf{\tiny{$\mathbf{MMD_{a}}$}} & 
\negthinspace\negthinspace\textbf{\tiny{$\mathbf{MMD_{s}}$}} \\
\cline{2-8}
\noalign{\smallskip}
\hline
\parbox[t]{1mm}{\multirow{3}{*}{\hspace{-0.21em}\rotatebox[origin=c]{90}{\footnotesize{\textbf{NTU-120}}}}}
& \multicolumn{1}{c|}{\cellcolor{gray!25}} & 
\multicolumn{3}{c|}{\cellcolor{gray!25}  \footnotesize{\emph{Cross-Subject} }}& 
\multicolumn{3}{c|}{\cellcolor{gray!25}  \footnotesize{\emph{Cross-Setup} }}\\
&\scriptsize{c-GAN \cite{mirza2014conditional}} & 
\negthinspace\negthinspace\scriptsize{$54.403$}   &  
\negthinspace\scriptsize{$1.037$} & 
\negthinspace\scriptsize{$1.104$}   & 
\negthinspace\negthinspace\scriptsize{$58.531$}   & 
\negthinspace\scriptsize{$1.082$}   & 
\negthinspace\scriptsize{$1.141$}   \\
&\scriptsize{\textbf{Kinetic-GAN}} & 
\negthickspace\scriptsize{$\mathbf{5.967}$}  & 
\negthickspace\scriptsize{$\mathbf{0.819}$}  & 
\negthickspace\scriptsize{$\mathbf{0.906}$}   & 
\negthickspace\scriptsize{$\mathbf{6.751}$}   & 
\negthickspace\scriptsize{$\mathbf{0.847}$}   & 
\negthickspace\scriptsize{$\mathbf{0.934}$}   \\
\hline

\end{tabular}
\vspace{-1em}
\end{center}
\caption{\textbf{Global movement generation results} on NTU-120 RGB+D with 120 different actions.}
\label{tab:ntu120}
\vspace{-1em}
\end{table}


\subsection{Local Movement Settings}
\label{sec:local}
Since most state-of-the-art methods were still limited to local movement settings, we also present the results under the same conditions for a fair comparison, where 3D locations were projected into 2D space and normalized positions. We follow the same settings as previous methods  \cite{habibie2017recurrent, wang2020learning,yu2020structure} and report the results on Table \ref{tab:local}. On both datasets, we verify the superiority of our model, where we achieve state-of-the-art performance with a large margin.

\newcolumntype{C}[1]{>{\centering\arraybackslash}p{#1}}
\newcolumntype{l}[1]{>{\arraybackslash}p{#1}}

\begin{table}[h]
\begin{center}
\begin{tabular}{|C{0.001\textwidth}|l{0.12\textwidth}|C{0.125\textwidth}|C{0.125\textwidth}|}
\cline{2-4}
\multicolumn{1}{c|}{} & 
\textbf{\footnotesize{Method}}  & 
\textbf{\tiny{$\mathbf{MMD_{a}}$}} & 
\textbf{\tiny{$\mathbf{MMD_{s}}$}} \\
\cline{2-4}
\noalign{\smallskip}
\hline
\parbox[t]{1mm}{\multirow{9}{*}{\hspace{-0.21em}\rotatebox[origin=c]{90}{\footnotesize{\textbf{Human$\mathbf{3.6}$M}}}}}
&\scriptsize{E2E \cite{villegas2018hierarchical}} & 
\scriptsize{$0.991$}   &  
\scriptsize{$0.805$} \\
&\scriptsize{EPVA \cite{villegas2018hierarchical}} & 
\scriptsize{$0.996$}   &  
\scriptsize{$0.806$} \\
&\scriptsize{adv-EPVA \cite{villegas2018hierarchical}} & 
\scriptsize{$0.977$}   &  
\scriptsize{$0.792$} \\
&\scriptsize{SkeletonVAE \cite{habibie2017recurrent}} & 
\scriptsize{$0.452$}  & 
\scriptsize{$0.467$}  \\
&\scriptsize{SkeletonGAN \cite{cai2018deep}} & 
\scriptsize{$0.419$}  & 
\scriptsize{$0.419$}  \\
&\scriptsize{c-SkeletonGAN \cite{wang2020learning}} & 
\scriptsize{$0.195$}   & 
\scriptsize{$0.218$} \\
&\scriptsize{c-GAN \cite{mirza2014conditional}} & 
\scriptsize{$0.161$}   & 
\scriptsize{$0.187$} \\
&\scriptsize{SA-GCN \cite{yu2020structure}} & 
\scriptsize{$0.146$}   &  %
\scriptsize{$0.134$} \\
&\scriptsize{\textbf{Kinetic-GAN}} & 
\scriptsize{$\mathbf{0.071}$} & 
\scriptsize{$\mathbf{0.082}$} \\\hline

\end{tabular}

\begin{tabular}{|C{0.001\textwidth}|l{0.12\textwidth}|C{0.05\textwidth}|C{0.05\textwidth}|C{0.05\textwidth}|C{0.05\textwidth}|}
\noalign{\vspace{0.5\smallskipamount}}
\cline{3-6}
\multicolumn{2}{c|}{} &  
\textbf{\tiny{$\mathbf{MMD_{a}}$}} & 
\textbf{\tiny{$\mathbf{MMD_{s}}$}}& 
\textbf{\tiny{$\mathbf{MMD_{a}}$}} & 
\textbf{\tiny{$\mathbf{MMD_{s}}$}} \\
\cline{3-6}
\noalign{\smallskip}
\hline
\parbox[t]{1mm}{\multirow{7}{*}{\hspace{-0.21em}\rotatebox[origin=c]{90}{\footnotesize{\textbf{NTU-2D RGB+D}}}}}
& \multicolumn{1}{c|}{\cellcolor{gray!25}} & 
\multicolumn{2}{c|}{ \cellcolor{gray!25}  \footnotesize{\emph{Cross-Subject} }}& 
\multicolumn{2}{c|}{ \cellcolor{gray!25}  \footnotesize{\emph{Cross-View} }}\\
&\scriptsize{SkeletonVAE \cite{habibie2017recurrent}} & 
\scriptsize{$0.992$}   &  
\scriptsize{$1.136$}  & 
\scriptsize{$1.079$}   &  
\scriptsize{$1.205$} \\
&\scriptsize{SkeletonGAN \cite{cai2018deep}} & 
\scriptsize{$0.698$}   &  
\scriptsize{$0.788$} & 
\scriptsize{$0.999$}   &  
\scriptsize{$1.311$}  \\
&\scriptsize{c-SkeletonGAN \cite{wang2020learning}} & 
\scriptsize{$0.338$}   &  
\scriptsize{$0.402$} & 
\scriptsize{$0.371$}   &  
\scriptsize{$0.398$}  \\
&\scriptsize{c-GAN \cite{mirza2014conditional}} & 
\scriptsize{$0.334$}   &  
\scriptsize{$0.354$}  & 
\scriptsize{$0.365$}   &  
\scriptsize{$0.373$} \\
&\scriptsize{SA-GCN \cite{yu2020structure}} & 
\scriptsize{$0.285$}   &  
\scriptsize{$0.299$}  & 
\scriptsize{$0.316$}   &  
\scriptsize{$0.335$} \\
&\scriptsize{\textbf{Kinetic-GAN}} & 
\scriptsize{$\mathbf{0.256}$}  & 
\scriptsize{$\mathbf{0.273}$}  & 
\scriptsize{$\mathbf{0.295}$}   &  
\scriptsize{$\mathbf{0.310}$}  \\\hline

\end{tabular}
\vspace{-1em}

\end{center}
\caption{\textbf{Local movement generation results}. The MMD scores (lower is better) between real and synthetic samples generated over Human$3.6$M and NTU-2D RGB+D.}
\label{tab:local}
\vspace{-1.7em}
\end{table}


\subsubsection{Increasing temporal action length}
\label{ssec:action_recon}
\begin{figure}[t]
  \begin{center}
  \includegraphics[width=8.4cm]{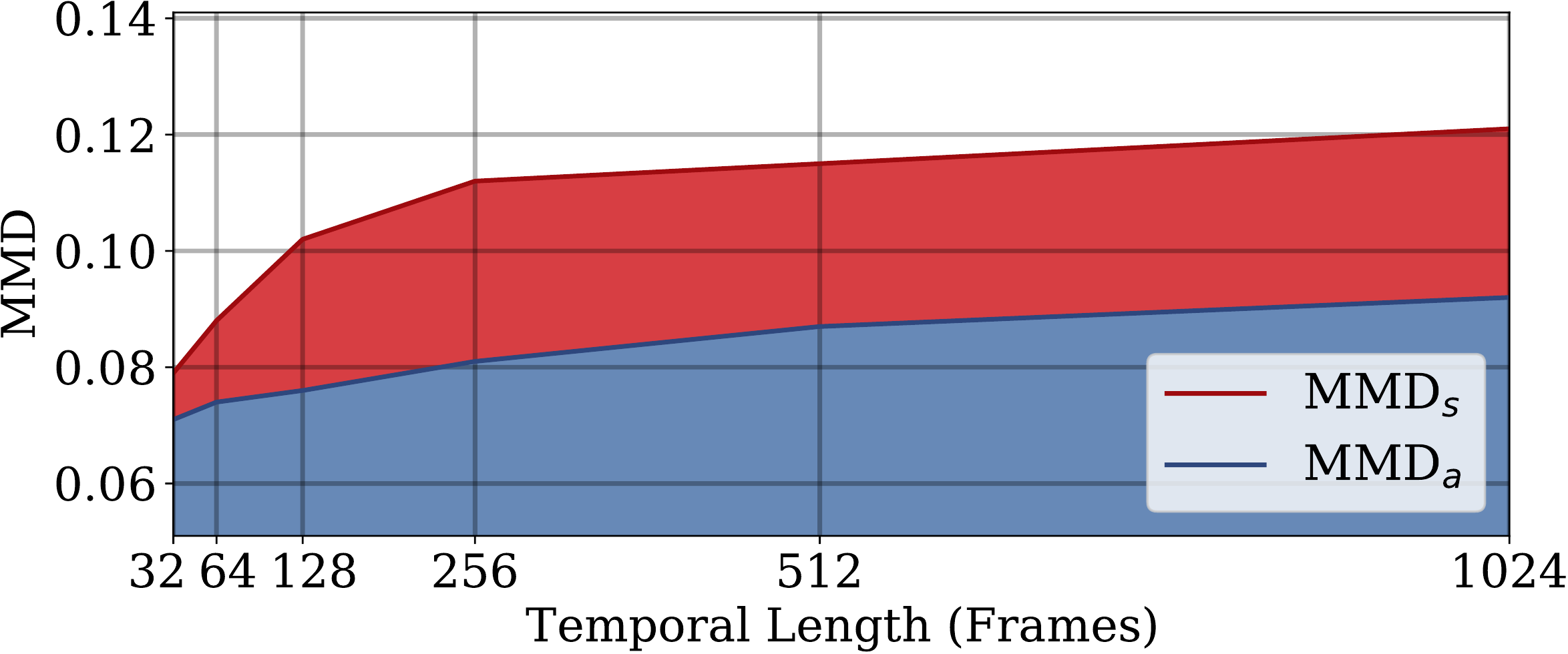}
  \caption{\textbf{Increasing action length} on Human$3.6$M.}
  \label{fig:time}
  \end{center}
\vspace{-1.8em}
\end{figure}

In addition to generation quality, action conditioning (desirable actions) and local/global movement, modelling long-term human actions is also a significant concept in action synthesis. However, NTU RGB+D could not be employed for such a study since the action execution average is 64 frames, where the remaining frames (maximum 300) are all set to 0 for normalization purposes. For this reason, such normalization can disturb the learning process of an action synthesis algorithm. So, we performed long-term experiments over Human$3.6$M, Fig.~\ref{fig:time}. Considering that our model can perform \textit{bidirectional temporal dependency}, mainly due to generating a whole human action sequence altogether, we can generate up to 1024 frames (34 seconds). Some autoregressive models \cite{fragkiadaki2015recurrent, zhou2018auto} gradually freeze poses during the sequence as a result of losing temporal dependencies for such sequence lengths. It can also be observed that its generation quality will stabilize over lengths $>256$ frames, which can also be positively regarded.
 
 \section{Conclusions and Further Work}
 \label{sec:Conclusions}
This paper introduced a novel Generative Adversarial Graph Convolutional Network for human action synthesis. By generating a human action in a holistic way directly from the latent space, we are able to better disentangle the variation factors through a mapping network, obtaining better representations in the latent space. Furthermore, the introduction of learnable noise injection modules facilitates the generation of variety without compromising the skeleton structure. As a result, we can generate up to 120 different complex actions, which, to the best of our knowledge, were particularly challenging for previous approaches under global movement settings. The proposed method was evaluated on three well known datasets (NTU RGB+D, NTU-120 RGB+D and Human$3.6$M), advancing the state-of-the-art performance metrics by a significant margin.

\footnotesize{\textbf{Acknowledgements}: This work was partially supported by the FCT/MEC through National Funds and by the FEDER-PT2020 Partnership Agreement under the Projects UIDB/50008/2020, POCI-01-0247-FEDER-033395, CENTRO-01-0247-FEDER-113023 - DeepNeuronic, operation Centro-01-0145-FEDER-000019 - C4 - Centro de Competências em Cloud Computing, co-funded by the European Regional Development Fund (ERDF) through the Programa Operacional Regional do Centro (Centro 2020), in the scope of the Sistema de Apoio a Investiga\c{c}\~{a}o Cient\'{i}fica e Tecnol\'{o}gica - Programas Integrados de IC\&DT and NOVA LINCS under grant ‘UIDB/04516/2020’. This research was also supported by ‘FCT - Funda\c{c}\~{a}o para a Ci\^{e}ncia e Tecnologia’ through the research grant ‘UI/BD/150765/2020’ and `2020.04588.BD'. }

\normalsize
\appendix
\section{Hyperparameters and training configurations}

\textbf{Datasets settings}. For global movement experiments in Section \ver{4.2}, which included NTU RGB+D \cite{shahroudy2016ntu} and NTU-120 RGB+D \cite{liu2019ntu} datasets, the temporal length of the skeleton sequences was normalized to $t=64$ frames. The reason behind the chosen temporal length resides in the action execution average of the dataset (64 frames). Despite both datasets containing some annotation errors (some inaccurate 3D joints position), no sample filtering was applied. We confirm the superiority of our method in action conditioning by using every action class in both datasets (60 for NTU RGB+D and 120 for NTU-120 RGB+D). For local movement experiments in Section \ver{4.3}, which included Human$3.6$M \cite{ionescu2013human3} and NTU-2D RGB+D \cite{shahroudy2016ntu} datasets,  the same settings were applied as previous approaches \cite{habibie2017recurrent, wang2020learning,yu2020structure}. Specifically, the temporal length was normalized to $t=50$ frames, the number of action classes used are 10, and both datasets were normalized from real/global movement to local movement, which facilitates the generation process. In Human$3.6$M \cite{ionescu2013human3} dataset, the following action classes are used: \textit{sitting, sitting down, discussion, walking, greeting, direction, phoning, eating, smoking and posing}. In NTU-2D RGB+D \cite{shahroudy2016ntu} dataset, the following action classes are used: \textit{drinking water, jump up, make phone call, hand waving, standing up, wear jacket, sitting down, throw, cross hand in front and kicking something}. Also, for a fair comparison, training samples from NTU-2D RGB+D \cite{shahroudy2016ntu} were carefully selected from each class on NTU RGB+D \cite{shahroudy2016ntu} similar to previous methods \cite{habibie2017recurrent, wang2020learning,yu2020structure}. \par

\textbf{Training configurations}. We train the networks using Adam \cite{kingma2014adam} optimizer with $\alpha = 2 \times 10^{-4}$, $\beta_1 = 0.5$, $\beta_2 = 0.999$ and $\epsilon = 10^{-8}$ for all datasets with a minibatch size of 32. Since we rely on the WGAN-GP loss \cite{gulrajani2017improved}, we set $n_{critic} = 5$, which sets the number of iterations of the discriminator per generator iteration.\par

\textbf{Upsampling and downsampling details}. As illustrated in Fig. \ver{3} (paper), the spatial resolution of the skeleton is increased from the intermediate latent point as $1 \rightarrow 5 \rightarrow 11 \rightarrow 25$ joints for the NTU RGB+D \cite{shahroudy2016ntu}, NTU-2D RGB+D \cite{shahroudy2016ntu} and NTU-120 RGB+D \cite{liu2019ntu} datasets. For the Human$3.6$M \cite{ionescu2013human3} dataset the spatial resolution is increased as $1 \rightarrow 2 \rightarrow 7 \rightarrow 15$ joints. In all datasets, the temporal resolution is increased by doubling $t/16$ until reaching the dataset's temporal length $t$. The same resolutions reversed are applied for the downsampling paths in the discriminator.\par

\textbf{Mapping network structure}. Our non-linear mapping network comprises fully connected layers with 512 as the dimensionality of the input and output activations. As demonstrated in Table \ver{2}, the increasing number of different subjects in the training data results in a more complex latent representation requiring a deeper mapping network. For this reason, we set 6 layers for the Human$3.6$M \cite{ionescu2013human3}, and 8 layers for the NTU-120 RGB+D \cite{liu2019ntu} dataset. NTU RGB+D and NTU-2D RGB+D \cite{shahroudy2016ntu} datasets follow the same settings as studied in Table \ver{2}.\par
\textbf{Noise injection details}. The noise injector described in Section \ver{3.4.1} samples a random noise $\bm{r}_l$ using $\mathcal{N}(0, 1)$. Each joint at resolution level $l$ has a respective weight to each channel and receives a different noise added channel-wise. This operation is applied to every generator's layer.

\section{Action complexity}

\begin{figure}[h]
  \begin{center}
  \includegraphics[width=8.4cm]{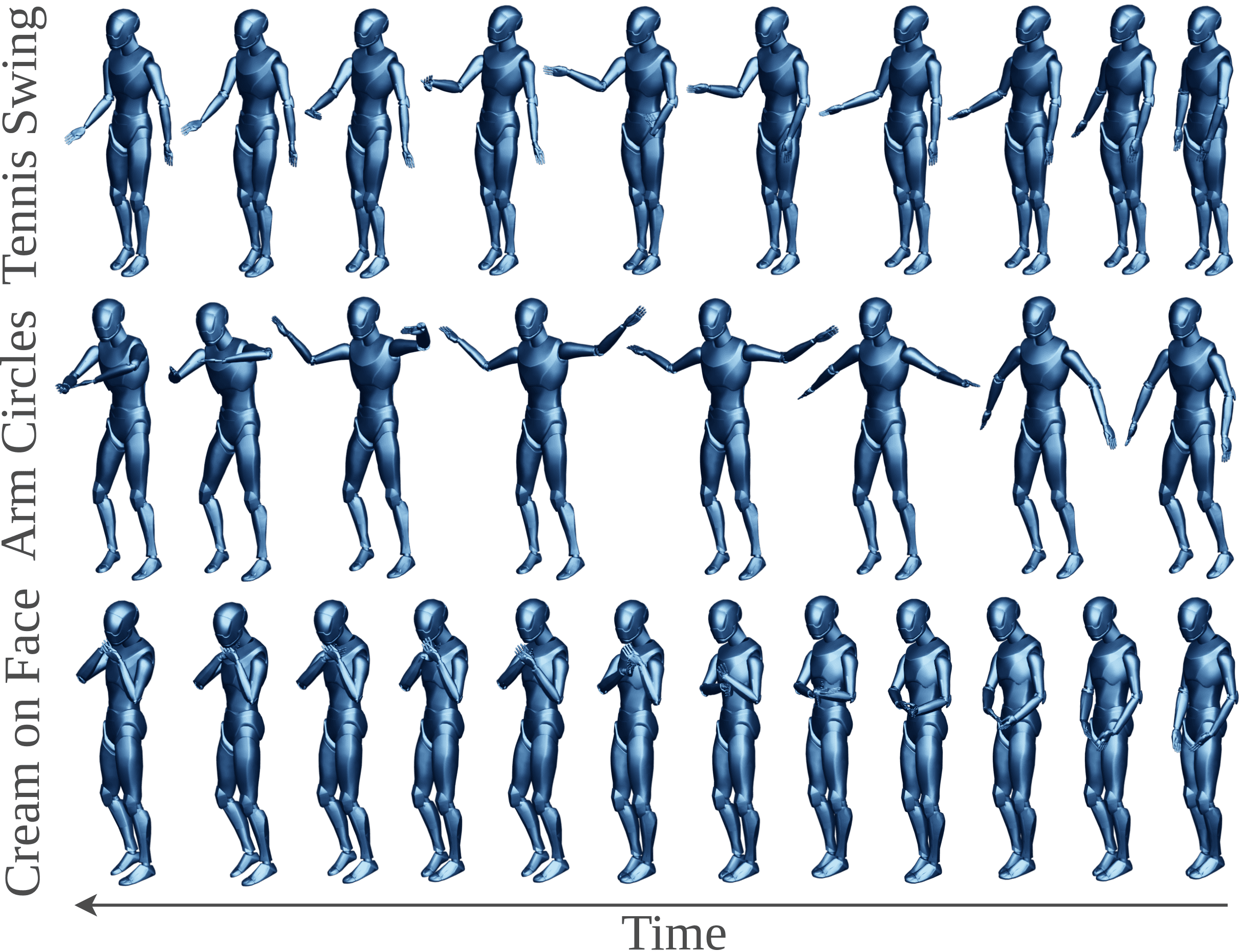}
  \caption{\textbf{Synthetic set of actions} generated by our graph convolutional generator trained on NTU-120 RGB+D \cite{liu2019ntu}.}
  \label{fig:ntu120}
  \end{center}
  \vspace{-1.8em}
\end{figure}

\begin{figure}[t!]
  \begin{center}
  \includegraphics[width=8.4cm]{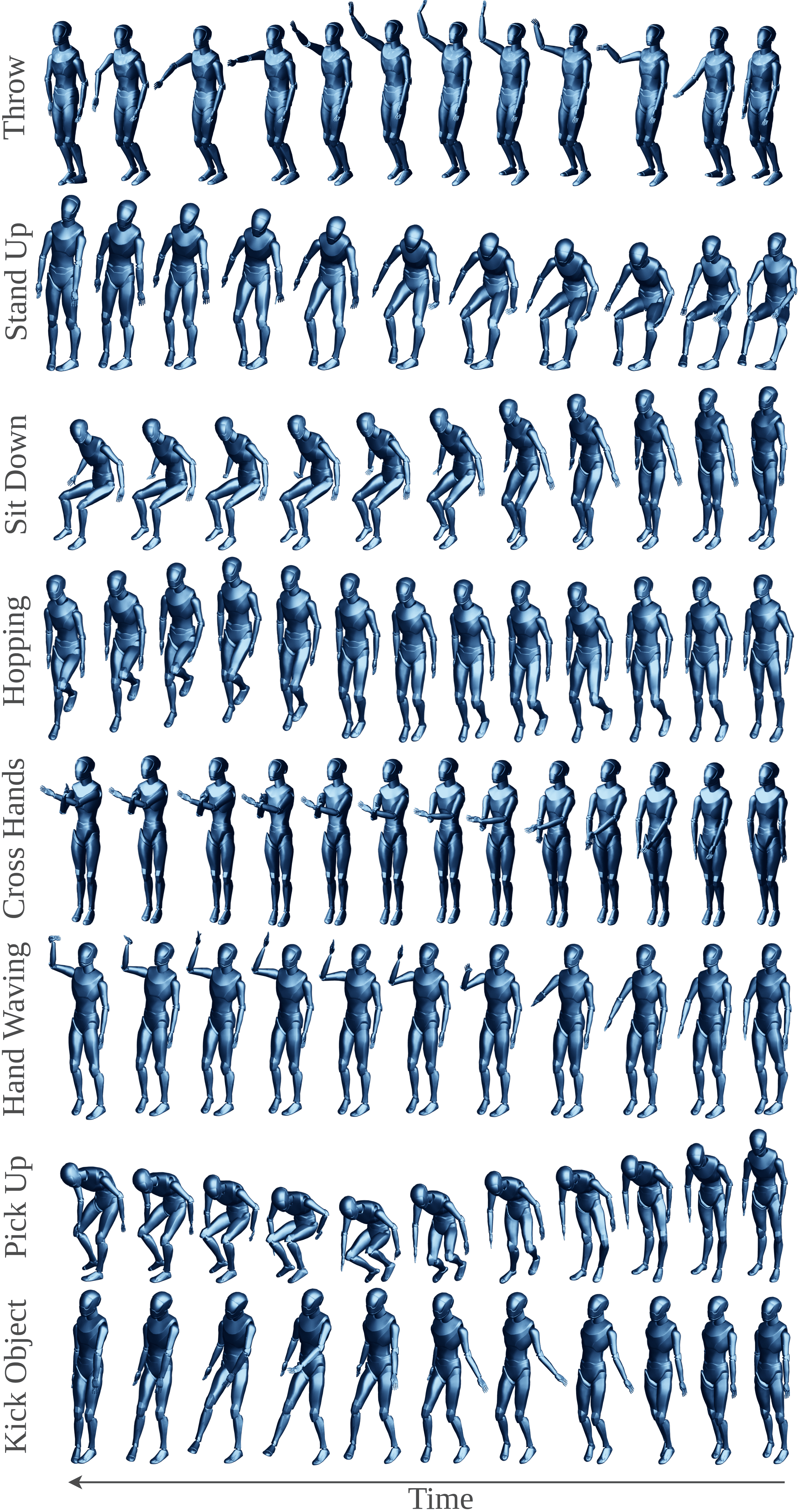}
  \caption{\textbf{Synthetic set of actions} generated by our graph convolutional generator trained on NTU RGB+D \cite{shahroudy2016ntu}.}
  \label{fig:ntu60}
  \end{center}
  \vspace{-1.8em}
\end{figure}

We include several action samples synthesised by our graph convolutional generator that demonstrate various aspects related to action complexity (see also accompanying video). Apart from the ability to generate up to 120 different action classes, we are able to generate global (real) body movement in 3D space, which, to the best of our knowledge, such complex actions under global movement settings had proven to be uncharted territory for previous methods. Figure \ref{fig:ntu60} shows different action examples illustrating the detail and expressiveness achievable using our method in NTU RGB+D \cite{shahroudy2016ntu}. In Figure \ref{fig:ntu120}, we demonstrate the ability to generate desired actions among 120 different classes from NTU-120 RGB+D \cite{liu2019ntu}.

{\small
\bibliographystyle{ieee_fullname}
\bibliography{egbib}
}

\end{document}